\documentclass{article}
\usepackage{PRIMEarxiv}

\usepackage[utf8]{inputenc} 
\usepackage[T1]{fontenc}    
\usepackage{url}            
\usepackage{booktabs}       
\usepackage{amsfonts}       
\usepackage{nicefrac}       
\usepackage{microtype}      
\usepackage{lipsum}
\usepackage{fancyhdr}       
\usepackage{graphicx}       

\usepackage{mathptmx}
\usepackage{mathrsfs}
\usepackage{mathtools}
\usepackage{amssymb}
\usepackage{epsfig}
\usepackage{multirow}
\usepackage{siunitx}
\usepackage{ifthen}
\usepackage{anyfontsize}
\usepackage{arydshln}

\usepackage[
  bookmarks=false,
  pdfpagelabels=false,
  hyperfootnotes=false,
  hyperindex=false,
  pageanchor=false,
  colorlinks,
]{hyperref}

\DeclareMathOperator*{\argmax}{arg\,max}

\author{
\begin{tabular}[t]{c@{\extracolsep{2.0em}}c@{\extracolsep{2.0em}}c@{\extracolsep{2.0em}}c}    
    \multicolumn{3}{c}{    
        \begin{tabular}{c}
        Panos Achlioptas${}^{1}$ \thanks{Corresponding author.}\\
        \fontsize{9}{9} \selectfont \texttt{panos@snap.com}
        \end{tabular}
        \begin{tabular}{c}
         Maks Ovsjanikov${}^{2}$ \\
         \fontsize{9}{9} \selectfont \texttt{maks@lix.polytechnique.fr}
        \end{tabular}
        \begin{tabular}{c}
        Leonidas Guibas${}^{3}$ \\
        \fontsize{9}{9}\selectfont \texttt{guibas@cs.stanford.edu}
        \end{tabular}
        \begin{tabular}{c}
        Sergey Tulyakov${}^{1}$ \\
        \fontsize{9}{9} \selectfont \texttt{stulyakov@snap.com}
        \end{tabular}}\vspace{9pt}\\
    \multicolumn{3}{c}{${}^{1}$Snap Inc.}\\
    \multicolumn{3}{c}{${}^{2}$LIX, Ecole Polytechnique, IP Paris}\\
    \multicolumn{3}{c}{${}^{3}$Stanford University}\\
\end{tabular}
}

\newcommand{\paper}{main}

\begin{document}

\pagestyle{fancy}
\thispagestyle{empty}
\rhead{\textit{Achlioptas et al.} Affection} 

\ifthenelse{\equal{\paper}{main}}{
\title{
  
    Affection: Learning Affective Explanations for Real-World Visual Data

}}

\ifthenelse{\equal{\paper}{supp}}{
\title{
Affection: Learning Affective Explanations for Real-World Visual Data \\Supplemental Material
}}

\date{}
\maketitle
\newcommand{\datasetName}{Affection}
\newcommand{\datasetNumberOfImages}{85,007}
\newcommand{\datasetNumberOfUtter}{526,749}
\newcommand{\datasetNumberOfAnnotators}{6,283}
\newcommand{\datasetNumberOfHours}{15,607}

\newcommand{\datasetPrcOfPositiveEmotions}{71.3\%} 
\newcommand{\datasetPrcOfNegativeEmotions}{21.1\%}
\newcommand{\datasetPrcOfSEEmotions}{7.6\%}

\newcommand{\datasetImgWithBothNegPosEmotions}{50.0\%}
\newcommand{\datasetImgWithNegPosSEEmotions}{66.0\%}

\newcommand{\datasetPrcOfStrongEmoMajority}{67.5\%} 
\newcommand{\datasetABSOfStrongEmoMajority}{57,381}

\newcommand{\datasetAvgWordsPerUtter}{18.8}
\newcommand{\datasetNumberOfTokens}{41,275} 




\newcommand{\datasetPrcOfNeutralUtters}{10.5\%} 


\newcommand{\datasetPrcOfSimiles}{19.7\%}
\newcommand{\allImagesConsidered}{244,172}
\newcommand{\numTestUtterances}{52,188}
\newcommand{\resnetD}{34}
\newcommand{\webpage}{\url{https://affective-explanations.org}}

\def\eg{e.g.,}
\def\ie{i.e.,}
\newcommand{\NameOfAppendix}{Supp.~Mat.}
\newcommand{\mypara}[1]{\vspace*{-10pt}\paragraph{#1}}
\newcommand{\nospacepara}[1]{\noindent{\bf {#1}}}
\newcommand{\flickr}{Flickr30k Ent.}
\newcommand{\coco}{COCO}
\newcommand{\cococite}{\cite{lin14eccv,coco_chen2015}}
\newcommand{\vg}{Visual Genome}

\newcommand{\red}[1]{{\color{red}{#1}}}
\newcommand{\yellow}[1]{{\color{yellow}{#1}}}
\newcommand{\blue}[1]{{\color{blue}{#1}}}
\newcommand{\qq}[2]{\red{{#1}}\blue{#2}} 
\newcommand{\panos}[1]{{\textcolor{blue}{Panos: #1 $\blacksquare$}}}
\newcommand{\sergey}[1]{{\textcolor{green}{Sergey: #1 $\blacksquare$}}}
\newcommand{\leo}[1]{{\textcolor{cyan}{Leo: #1 $\blacksquare$}}}
\newcommand{\maks}[1]{{\textcolor{red}{Maks: #1 $\blacksquare$}}}

\newcommand{\cg}[1]{{\textcolor[RGB]{102,178,84}{#1}}}

\ifthenelse{\equal{\paper}{main}}{

\begin{abstract}
In this work, we explore the emotional reactions that real-world images tend to induce by using \textit{natural language} as the medium to express the \textit{rationale} behind an affective response to a given visual stimulus. To embark on this journey, we introduce and share with the research community a large-scale dataset that contains emotional reactions and free-form textual explanations for \datasetNumberOfImages{} publicly available images, analyzed by \datasetNumberOfAnnotators~annotators who were asked to indicate and explain \textit{how and why} they felt in a particular way when observing a particular image, producing a total of \datasetNumberOfUtter{} responses. Even though emotional reactions are subjective and sensitive to context (personal mood, social status, past experiences) – we show that there is significant common ground to capture potentially plausible emotional responses with a large support in the subject population. In light of this key observation, we ask the following questions: i) Can we develop multi-modal neural networks that provide reasonable affective responses to real-world visual data, explained with language? ii) Can we steer such methods towards producing explanations with varying degrees of pragmatic language or justifying different emotional reactions while adapting to the underlying visual stimulus? Finally, iii) How can we evaluate the performance of such methods for this novel task? With this work, we take the first steps in addressing all of these questions, thus paving the way for richer, more human-centric, and emotionally-aware image analysis systems. Our introduced dataset and all developed methods are available on \webpage{}.
\end{abstract}


\section{Introduction}
\label{sec:introduction}

One can argue that a central goal of computer vision systems has been to gain a semantic, deep, cognitive, etc. \textit{understanding} of visual stimuli~\cite{deep_l_survey,CHAI2021100134}. But what exactly do we mean by this understanding? The vast majority of existing image analysis systems focus solely on image \textit{content}~\cite{CHAI2021100134}. Such content can most often be summarized by either listing the objects the image depicts or performing scene-level analysis to capture, and often express in language, relations between shown objects or associated activities  (e.g., ``a person walking along a beach''). Although models aimed at image analysis and captioning have achieved unprecedented success during the past years~\cite{Sharma_2021,Stefanini2022FromST}, they largely ignore the more subtle and complex \textit{interactions} that might exist between the image and its potential viewer.

In this work, our primary goal is to take a step toward a more viewer-centered understanding going \textit{beyond} factual image analysis by incorporating the \textit{effect} that an image might have on a viewer. To capture this effect, we argue that \textbf{emotional responses} provide a \textbf{fundamental link }between the visual world and human experience. We thus aim to understand what kinds of emotions a given image can elicit to different viewers and, most importantly, \textit{why?}.

Emotion perception and recognition are influenced by and integrate many factors, from neurophysiological to cultural, from previous subjective experiences to social and even political context~\cite{Phylogeny_emotions}. Thus, capturing and potentially reproducing plausible emotional responses to visual stimuli is significantly more challenging than standard image analysis, as it also involves an inherently \textit{subjective} perspective, which is at the core of perception and consciousness~\cite{dehaene2014consciousness}. 

We argue that image captioning and reasoning systems should start addressing the dimension of subjectivity more systematically, following studies like those regarding consciousness~\cite{dehaene2014consciousness}, personality~\cite{shuster2018engaging} or emotions~\cite{barrett2017emotions}. Furthermore, as we demonstrate below, an affective-oriented analysis offers a new and vibrant source of information about both the contents of the image (its emotionally salient features, the interaction between an image and an audience) and the photographer/artist (why was the given image taken in the first place).

To proceed with the goal of establishing a novel approach to affective analysis of real-world images, we leverage the fact that while emotions are not themselves linguistic concepts, one of the most readily \textit{accessible} expressions we have for them is through language~\cite{OrtonyBook}. Specifically, inspired by recent advances in affective captioning of art-works~\cite{achlioptas2021artemis}, we study emotional responses catalyzed by real-world visual data in conjunction with human/subject-induced \textit{explanations}. This approach links emotions with linguistic constructs, which crucially are easier to curate \textit{at scale} compared to other media (e.g., fMRI scans). Put together, our work expands on the recent effort of Achlioptas \textit{et al.}~\cite{achlioptas2021artemis} by considering a visio-linguistic and emotion analysis across a large set of \textit{real-world images}, not only restricted to visual art.
 
Our main contributions to this end are two-fold: first, we curate a large-scale collection of \datasetNumberOfUtter{}~\textit{explanations justifying emotions} experienced at the sight of \datasetNumberOfImages~different real-world images selected from five public datasets. The collected explanations are given by \datasetNumberOfAnnotators~annotators spanning many different opinions, personalities, and tastes. The resulting dataset, which we term \textbf{\textit{Affection}}, turns out to be very rich in visual and linguistic variations, capturing a wide variety of both the underlying real-world depicted phenomena and their emotional effect. Second, we perform a linguistic and emotion-centric analysis of the dataset and, most importantly, use it to produce deep neural listeners and speakers trained to comprehend, or generate plausible \textit{samples} of visually grounded explanations for emotional reactions to images. Despite the aforementioned subjectivity and thus the more challenging nature of these tasks compared to purely descriptive visio-linguistic tasks (e.g., COCO-based captioning~\cite{coco_chen2015}), our methods appear to learn \textit{common} biases of how people react emotionally, e.g., the presence of a shark is much more likely to raise fear than the presence of a peacefully sleeping dog. Such \textit{common sense} expectations are well captured in Affection, which is why we believe even black-box approaches like ours show promising results. 

Interestingly, while the same high-level principle is what enabled the established image-to-emotion \textit{classification} systems, in our case, by learning emotions through language, our resulting analysis is more nuanced and also comes with \textit{explanatory} power. E.g., a dog showing their teeth while snarling can also be fear producing since as \textit{explained} it might hurt someone (see Figure~\ref{fig:neural_productions_emo_grounded_pragmatic} for neural-based generations demonstrating this fact). 

Last, as we demonstrate in this work, even `naive/basic' neural speaking architectures,  when trained with Affection, learn to generalize in creating, on average, sensible affective explanations. However, to enable minimal control over the generated captions, we also explore two variants of trained affective neural captioning systems: an \textit{emotionally-grounded} one, which is pre-conditioned on a given emotion category~\cite{achlioptas2021artemis}, and a second, \textit{pragmatic} variant~\cite{GOODMAN2016818,andreas-klein-2016-reasoning,achlioptas2019shapeglot}. The latter variant allows some control on the level of factual visual details that are used when providing an explanation (e.g., `The sky looks beautiful'  to `The blue colors of the sky and the sea in this sunset make me happy'). Interestingly, we also demonstrate that the pragmatic variant can help to mitigate mode collapse to the simplest explanations and demonstrates richer and more diverse language across different images.

In summary, we make the following key contributions:
\begin{itemize}
    \item We introduce the task of \textit{\underline{A}ffective \underline{E}xplanation \underline{C}aptioning}, dubbed \textbf{AEC}, for real-world images.
    \item We curate and share with the community \textit{Affection}: a large-scale dataset capturing \datasetNumberOfUtter{} emotional reactions and corresponding explanations collected by \datasetNumberOfAnnotators{} annotators working \datasetNumberOfHours{} hours in building it.
    \item We exploit this dataset to tackle AEC, by combining and exploring the interplay of a variety of neural network components, including standalone emotion classifiers for text or images,
    affective neural listeners, and most importantly, neural speakers that utilize the aforementioned components to develop different degrees of pragmatic and emotional control over their generations.
    \item Last but not least, we highlight that Affection's explanations include substantial image-\textit{discriminative} references (as revealed with our neural listening studies); and that all our neural speakers show strong performance on \textit{emotional Turing tests}. I.e., humans find their generations $\sim$60\%-65\% of the time likely to be uttered by other humans and not machines.
\end{itemize}

\section{Related Works}
\label{sec:related_works}

\paragraph{Emotion representation \& learning.}
\label{para:related-work:emotions}

Two of the most widely adopted paradigms for representing emotions in existing literature are the discrete \textit{categorical} system~\cite{ekman_emotions}, and the \textit{continuous} 2D-dimensional Valence-Arousal (VA) model~\cite{emotion_rep, russell1980circumplex}. The former assumes a predefined (typically small) set of universally hardwired affective states, e.g., happiness, sadness, etc.; while the latter proposes that emotions arise from a combination of two fundamental dimensions: the  emotional arousal (strength or intensity of emotion), and the emotional valence (degree of pleasantness or unpleasantness of the emotion)~\cite{Hamann_emotions}. In our work and following previous studies~\cite{img_clf_art,Yanulevskaya-emotions,Zhao2014art,emotion_clf,achlioptas2021artemis}, we adopt the categorical system of emotion-representation, and use the same set of eight emotion categories, considered in those works. Specifically, we regard: \textit{anger}, \textit{disgust}, \textit{fear}, and \textit{sadness} as negative emotions, and \textit{amusement}, \textit{awe}, \textit{contentment}, and \textit{excitement} as positive ones. In line with previous works (\cite{Mikels_2005, emotion_clf, achlioptas2021artemis}), we treat \textit{awe} as a positive emotion when analyzing \datasetName. It is important to note that while we opt for the categorical system so as to stay closer to relevant existing works, there is still an open and very active debate regarding the nature of emotions and their optimal representation~\cite{barrett2017emotions,Hamann_emotions}.

\paragraph{Image captioning.} There is a vast amount of literature on the topic of image captioning~\cite{show-tell,xu2015show,vilbert,cornia2019show,cornia2020meshed,Luo2022AFS} (please see also review~\cite{Stefanini2022FromST} for a comprehensive exposition). Most of these works typically concern neural models trained and tested with \textit{descriptive} image-captions using some of the now well-established datasets~\cite{Kazemzadeh,coco_chen2015,conceptual-captions,VG_Krishna_2017,mao16,pont2020connecting}. Furthermore, relevant tasks such as VQA~\cite{vqa_original} , or works that lift image captioning~\cite{chen2021scan2cap,3d_Scent}, or reference disambiguation in 3D~\cite{zhenyu2019scanrefer,achlioptas2020referit_3d,goyal2020rel3d,SNARE,jain_3d}, still focus on descriptive (object- or scene-centric) language. A notable exception is the project called ArtEmis~\cite{achlioptas2021artemis}, which introduced a dataset and a series of tools for understanding and emulating the emotional effect of visual artworks. Similarly to that work, we focus on affective captions and develop affective neural speakers that aim to produce plausible textual utterances to capture the emotional effect of a given image. The key difference of our work compared to~\cite{achlioptas2021artemis} is that we focus on \textit{natural images}, not limited to art-works, making our dataset and tools of much broader scope and utility. It is also worth mentioning a connection of our work to recent developments at the intersection of NLP and Causal Representation Learning~\cite{causal_rep_and_nlp,causal_rep_learning}. Namely, our `captions' can be viewed as \textit{causal} explanations of an underlying observed phenomenon, that of an emotional reaction.

\paragraph{Pragmatics \& discriminative image analysis.} 
One of the increasingly prominent deep-learning tools is CLIP~\cite{CLIP}. CLIP is a discriminative visiolinguistic neural network that learns to assess the compatibility (likelihood of correspondence) between a caption and an arbitrary image. In line with recent approaches, we use it alongside our neural speakers, as well as in a standalone fashion to explore the discriminative properties of \datasetName. In particular, a discriminator like CLIP can provide indispensable guidance for generative models such as our neural speakers, by calibrating, or prioritizing their sampled productions (text), so as to increase the final caption-image compatibility~\cite{andreas-latent-lang,achlioptas_phd_thesis,dalle}. This process of re-ranking the captions to increase their relevance to the depicted image content, i.e., making them \textit{pragmatic}~\cite{andreas-klein-2016-reasoning,vedantam2017context,achlioptas2019shapeglot} is inspired by the Rational-Speech-Framework~\cite{GOODMAN2016818} and, as we show, can be particularly useful to mitigate mode-collapse and control the level of visual details expressed in our neural generations. To the best of our knowledge, our work is the first to apply CLIP in this manner for pragmatic captioning. Nevertheless, very recent works appear to use CLIP directly as an auxiliary loss function while training descriptive neural captioning systems~\cite{Cho2022CLIPReward}. Finally, it is worth mentioning here the work of Bondielli and Passaro~\cite{clip_for_img_emo_rec} who showed that CLIP can be used both in a zero-shot fashion, or by fine-tuning it for emotion-based image classification; and also the work of Wang \textit{et al.} which shows that CLIP is capable of assessing the quality and abstract perception of images in a zero-shot manner.

\section{\datasetName~Dataset}
\label{sec:dataset}

\begin{figure*}[ht!]
\includegraphics[width=\textwidth]{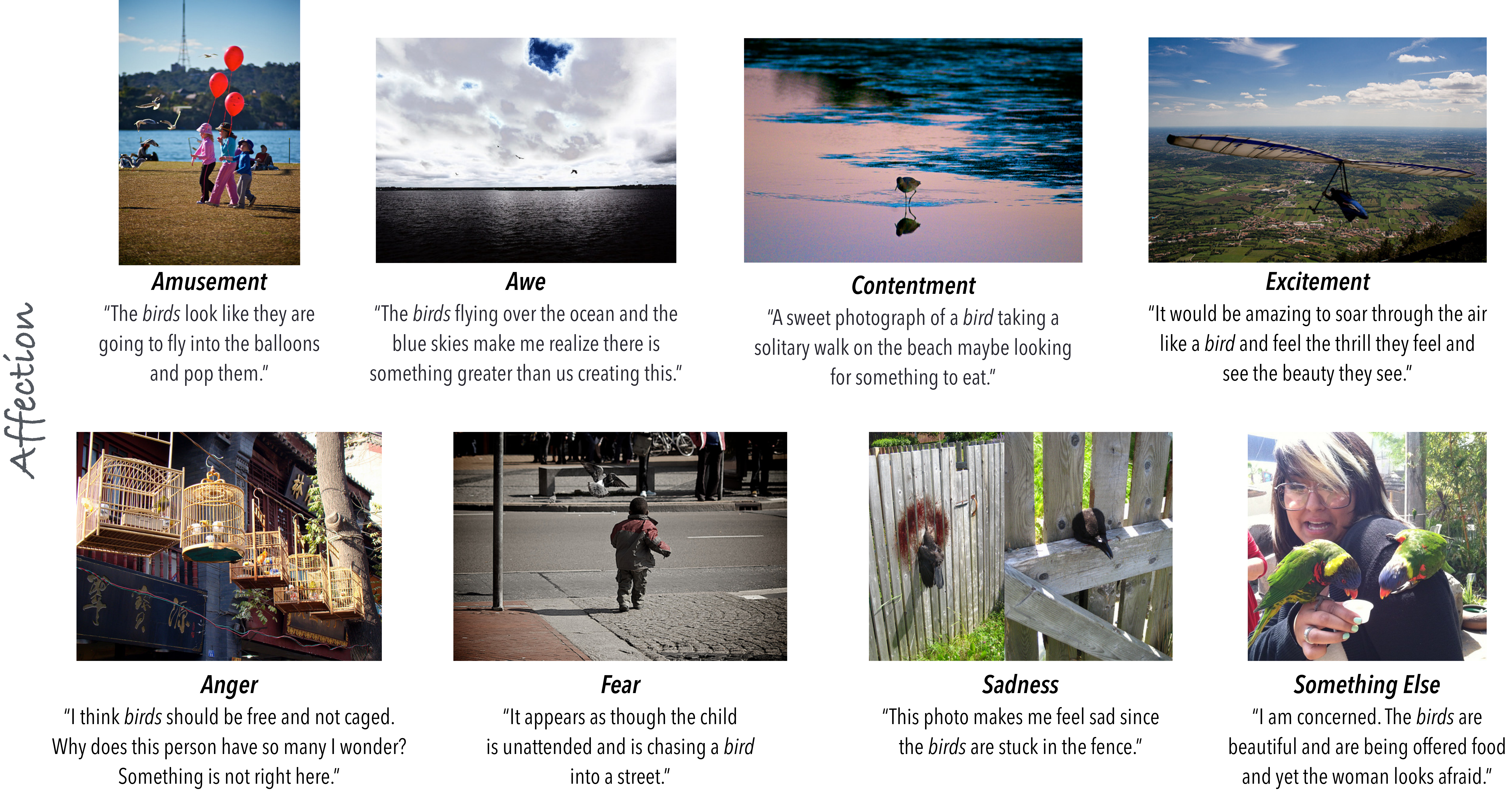}
\caption{{\bf Distinct emotional states and \textit{typical} explanations in Affection related to the entity of ``bird.''} As seen from these annotations in our Affection dataset, the collected explanations capture a wide range of abstract semantics and denote nuanced visio-linguistic associations between entities and the underlying explained emotion (shown in boldface). Noticeably, these semantics include common sense reasoning and 
cognitive-level understanding of an image, going \textit{beyond} recognizing its visible elements: \textit{birds flying near balloons can pop them}, \textit{when chased at a street can cause accidents}, \textit{when seen flying over vast landscapes, birds can provoke deeply existential thoughts.} The shown images are included in the datasets covered by Affection, no ownership or copyright claim is made by the authors of this article.}
\label{fig:dataset_teaser}
\end{figure*}

The \textit{\textbf{\datasetName}} (\underline{\textbf{Affect}}tive  Explanat\underline{\textbf{ion}}s)~dataset is built on top of images existing in the publicly available datasets MS-COCO~\cite{coco_chen2015}, Emotional-Machines~\cite{emotion_clf_valence_arousal}, Flickr30k Entities~\cite{plummer2015flickr30k}, Visual Genome~\cite{krishna2017visual}, and the images used in the image-to-emotion-classification work of Quanzeng \textit{et al.}~\cite{emotion_clf}. 

In total, we annotate \datasetNumberOfImages{} \textit{unique} images corresponding to a carefully curated subset of the \allImagesConsidered{} images contained in the above datasets. Specifically, we use \textit{all} images studied by Emotional-Machines and  Quanzeng \textit{et al.}, which have been specifically curated as to consistently evoke emotions among the corresponding annotators of these  works. Also, we further use these images and a ResNet-based~\cite{resnet} visual embedding pretrained on ImageNet~\cite{deng2009imagenet} to find their 3-Nearest-Neighbors in each of the remaining datasets (COCO, Visual-Genome, FlickR30k-Entities) when building Affection.

We annotate the aforementioned images by asking \textit{at least} \textbf{6} annotators per image to express their \textit{dominant} emotional reaction along with an explanation for their response. Specifically, we follow the setup used in the work of Achlioptas \textit{et al.}~\cite{achlioptas2021artemis}. Upon observing an image an annotator is asked first to indicate their dominant emotional reaction by selecting among the eight emotions mentioned in Section \ref{sec:related_works}, or a ninth option, listed as `something-else'. This latter option allows the annotators to indicate finer grained emotions not explicitly listed in our UI (see Figure~\ref{fig:dataset_teaser} for an example), or to explain why they might not have \emph{any} strong emotional reaction to the specific image (in total the annotators used this latter option \datasetPrcOfSEEmotions{} of the time). Upon completing the first step, each annotator is further asked to provide a textual explanation for their choice in natural language. The explanation should  include at least one specific reference to visual elements depicted in the image. See Figure~\ref{fig:dataset_teaser} for examples of \textit{typical} collected annotations. 

The resulting corpus consists of \textbf{\datasetNumberOfUtter{}} emotion indications and corresponding explanations, with the latter using a vocabulary of \datasetNumberOfTokens{} distinct tokens. We note that the \datasetNumberOfAnnotators{} annotators that worked to built \datasetName{} were recruited with Amazon's Mechanical Turk (AMT) services. For more details regarding Affection, our proposed methods and the experiments described below, we point the interested reader to the Supplemental Material~\cite{affection_supp}.

\begin{figure*}[ht]
    \centering
    \includegraphics[scale=0.32]{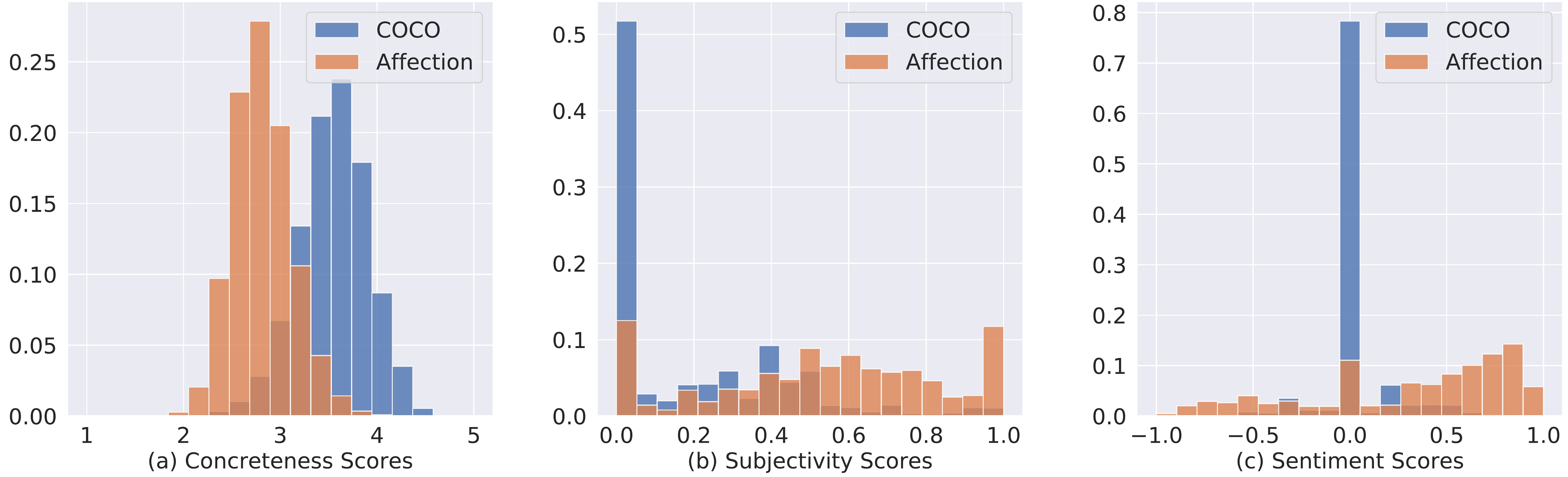}
    \caption{{\bf Characteristic properties of {\datasetName}}. Empirical distributions contrasting {\datasetName}'s explanations to the descriptive captions of COCO~\cite{coco_chen2015} along the axes of (a) \textit{Concreteness}, (b) \textit{Subjectivity}, and (c) \textit{Sentiment}.
  Evidently, the utterances of {\datasetName} are significantly more abstract, subjective, and sentimental.}
  \label{fig:analysis_teaser}
\end{figure*}

\subsection{Language-oriented Analysis}
\label{para:linguistic_analysis}

In the remainder of this section, we report the key characteristics of Affection, following the same overall approach in evaluating and analyzing the annotations as was used for the relevant affective dataset of Artemis~\cite{achlioptas2021artemis}. In the following sections, we describe our proposed learning-based methods and evaluation metrics, and conclude with more experiments.

\paragraph{Richness \& diversity.}
The average length of {\datasetName}'s explanations is \datasetAvgWordsPerUtter{} words. This is noticeably longer than the average length of utterances of ArtEmis and significantly longer than the captions of many other well-established and purely descriptive captioning datasets, as shown in Table~\ref{table:pos_per_captions}. Moreover, we use NLTK's part-of-speech tagger~\cite{bird2009natural}, to analyze {\datasetName}'s explanations in terms of their average number of contained nouns, pronouns, adjectives, verbs, and adpositions. Across \textit{all} these lexical categories, {\datasetName} contains a higher occurrence per caption, implying the use of a rich and complex vocabulary by its annotators. Also, when counting the \textit{unique} parts-of-speech that different annotators use in their explanations for the \textit{same} image  (Table~\ref{table:pos_per_image}), we find that Affection has a higher variety than other datasets. This last fact implies that our collected annotations aside of being lexically rich, they are also more \textit{diverse} across visual stimuli.

\begin{table}[ht]
    \centering
    \resizebox{\linewidth}{!}{%
    \begin{tabular}{lcccccc}
        \toprule
        Dataset & Words & Nouns & Pronouns & Adjectives & Adpositions & Verbs\\
        \midrule
        \textit{\datasetName} & \textbf{\datasetAvgWordsPerUtter} & \textbf{4.5} & \textbf{1.3} & \textbf{1.8} & \textbf{2.2} & \textbf{4.0}\\
        ArtEmis~\cite{achlioptas2021artemis} & 15.9 & 4.0 & 0.9 & 1.6 & 1.9 & 3.0 \\
        \flickr{}~\cite{young14tacl}  & 12.3 & 4.2 & 0.2 & 1.1 & 1.9 & 1.8 \\
        \coco{}~\cite{coco_chen2015} &  10.5 & 3.7 & 0.1 & 0.8 & 1.7 & 1.2 \\
        Conceptual Capt.~\cite{conceptual-captions}  & 9.6 & 3.8 & 0.2 & 0.9 & 1.6 & 1.1\\
        Google Refexp~\cite{mao16} & 8.4 & 3.0 & 0.1 & 1.0 & 1.2 & 0.8 \\
        \bottomrule
    \end{tabular}}\\[2mm]
    \caption{\small\textbf{Lexical comparison over distinct part-of-speech categories, per 
    \textit{individual captions}}.
    The average occurrences for the shown categories are significantly higher in \datasetName{}, indicating that it is comprised of a lexically \textbf{richer} and more complex corpus.}
    \label{table:pos_per_captions}
\end{table}
\begin{table}[ht]
    \centering
    \resizebox{\linewidth}{!}{%
    \begin{tabular}{l@{\hspace{5mm}}c@{\hspace{4mm}}c@{\hspace{4mm}}c@{\hspace{4mm}}c@{\hspace{4mm}}c}
        \toprule
        Dataset & Nouns & Pronouns & Adjectives & Adpositions & Verbs\\
        \midrule
        \textit{\datasetName} & \textbf{20.9} (3.4) & \textbf{4.4} (0.7) & \textbf{9.6} (1.5) & \textbf{8.6} (1.3) & \textbf{18.7} (3.0) \\
        ArtEmis~\cite{achlioptas2021artemis} & 18.7 (3.4) & 3.1 (0.6) & 8.3 (1.5) & 6.5 (1.2) & 13.4 (2.4) \\
        \flickr{}~\cite{young14tacl} & 12.9 (2.6) & 0.8 (0.2) & 4.0 (0.8) & 4.9 (1.0) & 6.4 (1.3) \\
        \coco{}~\cite{coco_chen2015} & 10.8 (2.2) & 0.6 (0.1) & 3.3 (0.7) & 4.5 (0.9) & 4.5 (0.9) \\
        Conceptual Capt.~\cite{conceptual-captions} & 3.8 (3.8) & 0.2 (0.2) & 0.9 (0.9) & 1.6 (1.6)  & 1.1 (1.1) \\
        Google Refexp~\cite{mao16} & 7.8 (2.2) & 0.4 (0.1) & 2.8 (0.8) & 2.9 (0.8) & 2.3 (0.6) \\
        \bottomrule
    \end{tabular}}\\[2mm]
    \caption{\small\textbf{Lexical comparison over distinct part-of-speech categories, per 
    \textit{individual images}}.
    The shown numbers indicate unique words per category averaged over individual images.
    In parentheses, we include a normalized version accounting for discrepancies in the number of annotators individual images might have. Evidently, \datasetName{}'s language is lexically more \textbf{diverse}.}
    \label{table:pos_per_image}
\end{table}

\paragraph{Abstractness \& subjectivity.}
The next two more idiosyncratic axes of language usage that we explore in connection to {\datasetName} are those of abstractness and subjectivity. Specifically, to measure the degree of abstractness (vs.~concreteness) of our corpus, we use the lexicon of Brysbaert \textit{et al.}~\cite{40k_absrtact_words} which provides for thousand word lemmas a scalar value (from 1 to 5) to indicate their concreteness. As an example, the words \textit{bird} and \textit{cage} represent fully concrete (tangible) entities, getting a score of 5, but the words \textit{freedom} and \textit{carefree} are considered more abstract concepts (with scores 2.34 and 1.88, resp.). On average, a uniformly random word of {\datasetName} scores 2.82 in concreteness while a random word of COCO does 3.55 (see also Figure~\ref{fig:analysis_teaser} (a)). In other words, the annotators of {\datasetName} make use of significantly more abstract concepts than in COCO. 
To further evaluate the degree to which {\datasetName} contains subjective language, we use the algorithm provided by TextBlob~\cite{textblob} which provides an estimate of how subjective an arbitrary sentence is, by assign to it a scalar value $\in[0,1]$. If a sentence contains no subjective references it is assigned a score of 0 (e.g., \textit{`The car is \underline{red}'}), while a sentence containing only subjective associations is assigned a score of 1 (e.g., \textit{`The car is \underline{nice}'}). As seen in the empirical distribution comparing our explanations to COCO's captions in terms of their subjectivity (Figure~\ref{fig:analysis_teaser} (b)), evidently Affection contains much more frequently subjective references.

\paragraph{Sentiment analysis.}
\label{para:dataset-sent-analysis}
Perhaps as expected by its nature, {\datasetName} also contains language that is highly sentimental. To measure this aspect of language usage, we use
a rule-based sentiment classifier (VADER~\cite{VADER}). VADER classifies \datasetPrcOfNeutralUtters{} of {\datasetName}'s explanations to the neutral sentiment, while it finds a significantly higher fraction ($77.4\%$) of the descriptive COCO-captions to fall in this category. In Figure~\ref{fig:analysis_teaser} ~(c) we present the histogram of VADER's estimated valences for the utterances contained in these two datasets. Valences closer to the extremes of the distribution (-1, 1) indicate highly negative or positive sentiment, respectively. Valences with small magnitude (closer to 0) indicate a neutral sentiment.

\begin{figure}[ht]
    \centering
  \includegraphics[ scale=0.5]{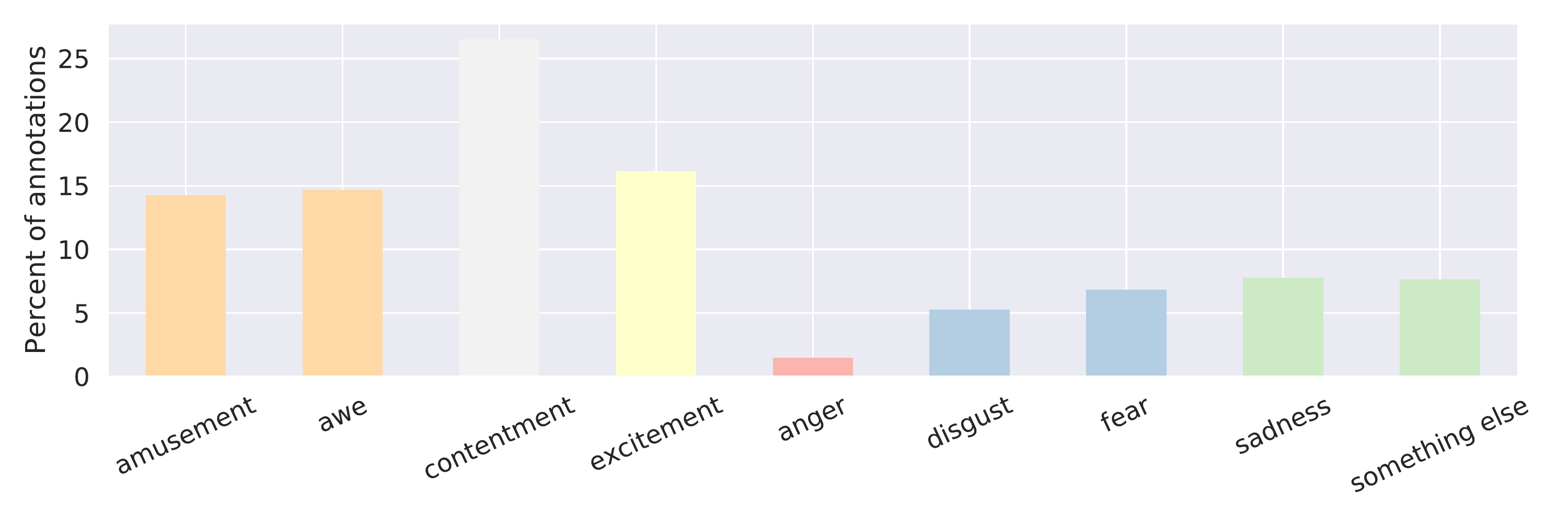}
  \vspace{-2mm}
  \caption{{\bf Empirical distribution of indicated dominant emotion categories accompanying {\datasetName}'s explanations}. Per our convention, the left-most four bars include \textit{positive} emotions, which are selected 
  \datasetPrcOfPositiveEmotions{} of the time. \textit{Negative} emotions (5th-8th bars)  appear much less frequently (\datasetPrcOfNegativeEmotions{}), while the  `something-else' is preferred \datasetPrcOfSEEmotions{} of the time.\vspace{2mm}}
  \label{fig:histogram_emotions_clicks}
\end{figure}

\paragraph{Comparing Affection to ArtEmis.}
The two previous analyses contrast our dataset with the descriptive captions of COCO. Another dataset that is more similar in nature with Affection is ArtEmis \cite{achlioptas2021artemis}. We note that Affection's language is similar in terms of abstractness to ArtEmis (average scores of 2.82 vs. 2.81). At the same time, Affection's language is more sentimental (VADER's classifier assigns \datasetPrcOfNeutralUtters{} vs. 16.5\% of each corpus to the neutral category), and more subjective (average subjectivity scores are 0.53 vs. 0.47, respectively). Please see the \NameOfAppendix{} for more detailed ArtEmis-based comparisons like those of Figure~\ref{fig:analysis_teaser}.

\subsection{Emotion-oriented Analysis.}
\label{para:emotion-analysis}

The annotators of Affection indicated a variety of dominant emotional reactions upon observing different visual stimuli. As seen in Figure~\ref{fig:histogram_emotions_clicks} positive emotions (see Section~\ref{para:related-work:emotions} for the used convention) were $\sim$3.4 more likely to occur than negative ones (\datasetPrcOfPositiveEmotions{} vs. \datasetPrcOfNegativeEmotions{}). Also, the ``something-else'' category was preferred  \datasetPrcOfSEEmotions{} of the time, and it included a large variety of subtler emotional reactions (e.g., curiosity, nostalgia, etc.). Despite the prevalence  of positive emotions overall, we note that crucially \textbf{{\datasetImgWithBothNegPosEmotions}} of images were annotated with at \textit{least one positive and one negative emotion}. While this result highlights the high degree of subjectivity inherent to our task, the next fact establishes that there is also \textit{significant} agreement among the annotators w.r.t.~their emotional reactions. 
Namely, {\datasetPrcOfStrongEmoMajority} ({\datasetABSOfStrongEmoMajority}) of the annotated images have a strong majority among their annotators who indicated the same fine-grained emotion. First, it is worth
mentioning that this fact establishes \datasetName{} also as one of the \textit{largest} publicly available image-to-emotion classification datasets, by merely concentrating on this \datasetABSOfStrongEmoMajority{} images and associating them only with their underlying majority label (i.e., following a similar strategy as the one used in the image-to-emotion-classification work of Quanzeng \textit{et al.}~\cite{emotion_clf}.). Second,
it is worth comparing our annotators' agreement with that of ArtEmis. Despite the fact that both datasets use the exact same protocol for annotation purposes, ArtEmis has much less agreement among its annotators: specifically, only 45.6\% of its annotated artworks attain a strong majority agreement. We hypothesize that this salient ($>20\%$) discrepancy is related to visual art being in general more ambiguous and thus evoking more polarizing emotions, compared to those resulting from the more familiar and well-understood experiences depicted or inspired by real-world images.

\paragraph{Remark on Joint Data Exploitation.} 
Affection is built on top of images for which rich annotations exist that are complimentary to our key affective captioning task. For instance, the images from Emotional-Machines~\cite{emotion_clf_valence_arousal} contain Valence-Arousal measurements (see Paragraph~\ref{para:related-work:emotions}), and the images from Quanzeng \textit{et al.} contain extra image/emotion-classification labels. Most importantly, for FlickR30K, Visual Genome, and COCO, descriptive captions accompany each annotated image of Affection. We believe that a joint exploitation of those annotations with the data in Affection offers many promising future directions, e.g., one can imagine neural speakers  that disentangle and control the `objective' parts of our visually grounded explanations, from their more subjective/personal references.

\section{Affective Tasks in Computer Vision}
\label{sec:method}

As mentioned above, Affection is a large scale dataset rich in visual and linguistic variations, containing significant amounts of sentimental, subjective and abstract language usage, grounded in natural images. In this section, and inspired by Affection's unique properties we introduce a series of problems and methods that make use of it. First we describe two straightforward emotion-oriented classification tasks, which we further exploit in our later designs. Section \ref{subsection:neural-listeners-and-speakers} then describes our main tools for neural listening and speaking, whose primary goal is to evaluate or produce plausible emotion explanations grounded in a visual input. While the methods we propose attain promising results, we emphasize that the value of Affection goes beyond the specific design of the neural listening/captioning systems that we evaluate. We thus consider the architectures used and evaluated below, as a preliminary validation of the value and utility of our dataset, while inviting the research community to build upon our observations and results.

\begin{figure*}[htbp]
\centering
\includegraphics[width=\textwidth]{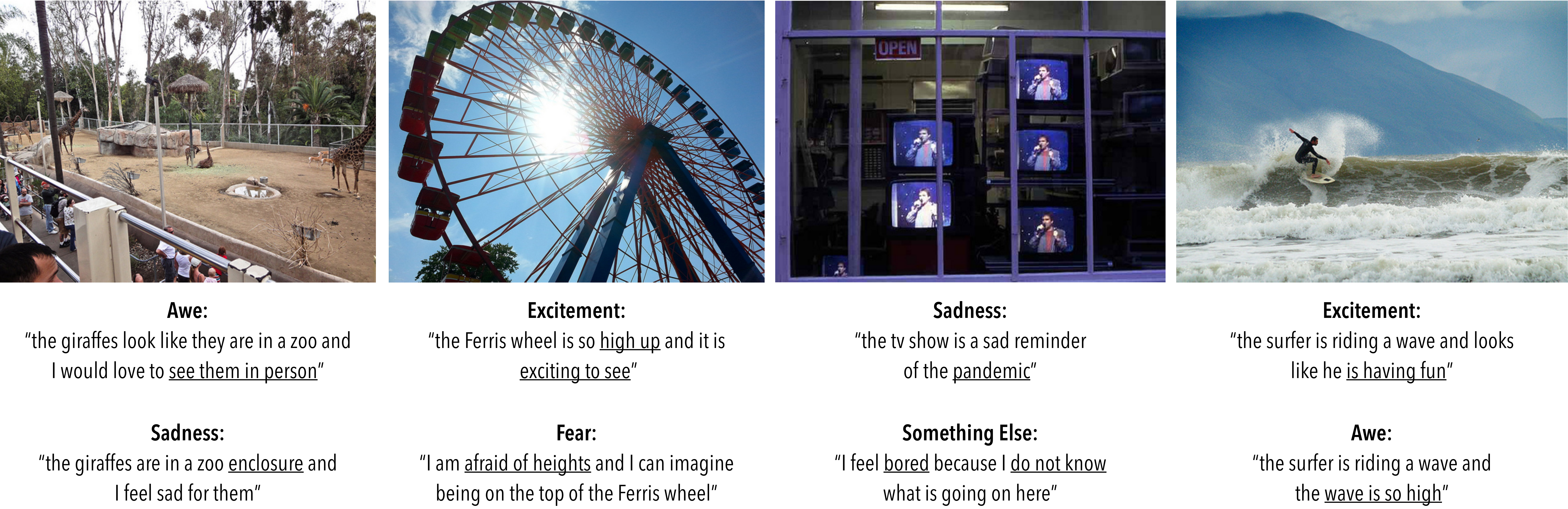}
\caption{\textbf{Curated examples of neural speaker generations with the default \textit{emotion-grounded} speaker variant on \textit{unseen} test images} The grounding emotion (shown in boldface fonts) is predicted during inference time by a separately trained image-to-emotion classifier. We ground the speaker's generation with two emotions for each image, corresponding to the most likely (top row) and second most likely (bottom row) predictions. As show in the figure, this variant provides a certain control over the output by aligning it to the requested/input emotion. The shown images are included in the datasets covered by Affection, no ownership or copyright claim is made by the authors of this article.}
\label{fig:neural_productions_with_emo_grounding}
\end{figure*}

\subsection{Basic Classification Tasks}
\label{subsection:aux_classifiers}
In designing our neural speakers we exploit solutions to two simple \textit{classification} problems, which we describe below. Our first problem consists in predicting the emotion class (among the 9 possibilities), that the given \textit{textual utterance} in Affection is associated with. Similarly to \cite{achlioptas2021artemis}, we formulate this problem as a standard 9-way \textit{text classification} problem. In our implementation we use an LSTM~\cite{lstm}-based text classifier trained from scratch using the standard cross-entropy loss. We also consider fine-tuning a pretrained BERT model~\cite{devlin2018bert} for this task. The second problem we explore consists of predicting the expected distribution of emotional reactions for a given \textit{image}. To this end, we fine-tune a ResNet-101 encoder~\cite{he2016deep} pretrained on ImageNet~\cite{deng2009imagenet} using, as a loss, the KL-divergence between the empirical emotional responses in Affection and the output predictions of the network. As we show below, these two classifiers (i.e., text to emotion and image to emotion), which we denote as $C_{emotion|text}$ and $C_{emotion|image}$ are instrumental in our design of neural speakers. Namely, they will allow us to both evaluate, as well as to control, the emotional content of the trained neural speakers (Sections~\ref{section:evaluation} and~\ref{subsection:emotion_grounded_speaker}). We note, however, that the two problems mentioned above have independent interest and we explore them in Section~\ref{sec:experimental_results}.

\subsection{Neural Listeners and Speakers }
\label{subsection:neural-listeners-and-speakers}

\paragraph{Neural comprehension (listening) with affective explanations.}\label{subsection:paragraph:neural-listeners}

To test the degree to which the explanations of Affection can be used to identify their underlying described image against random `distracting' images, we deploy two neural listeners~\cite{referit,GOODMAN2016818,achlioptas_phd_thesis}. First, we use Affection to train jointly and from scratch, an LSTM-based language encoder and a ResNet-based visual encoder under a self-contrastive criterion. Namely, during training, given a random batch of encoded image-caption pairs we optimize a cross-entropy loss that aligns the two modalities in a joint visual/language embedding space~\cite{dai2017contrastive,CLIP,ALIGN}. During training the application of such a loss effectively pulls together in the joint embedding space ground-truth pairs, while it pushes apart non-matched image-text pairs (within the same batch). Second, we deploy a pre-trained CLIP model~\cite{CLIP}, \textit{without} fine-tuning with Affection. Using a non-finetuned version of CLIP allows us to test this popular network \textit{as is} on our data, and to compare its performance against other datasets (e.g., COCO) in a straightforward and uniform way. During inference we input to the listeners a test image paired with a varying number of uniformly random distracting \textit{test} images, along with its ground-truth explanation, and output the image with maximal alignment (expressed in logits) to it. We note that to the best of our knowledge, the above described comprehension/listening studies are the first that address the extent to which \textit{affective language is also referential}~\cite{ref_model}.

\paragraph{Default speaker backbone} 

We use and adapt as our backbone model throughout our neural-speaker studies the Show-Attend-and-Tell (SAT)~\cite{xu2015show} model, due to its simplicity, solid performance and still wide usage in practice~\cite{defense_grid_features,rel_reasoning_survey}. Crucially, note that our adaptations for the neural speakers (i.e., using emotion-grounding, and pragmatic-inference) are \textit{generic and agnostic} to the choice of the base model~\cite{achlioptas2019shapeglot, achlioptas2021artemis}. 
The main element of our chosen backbone is an LSTM cell~\cite{lstm} which is grounded with the input image and which by using Affection's data learns to generate utterances that explain plausible emotional reactions to it. Specifically, at each time step the model learns to attend~\cite{attention_captioning} to different parts of the image (which is encoded by a separate ResNet-101 network), and by combining the `current' input token with the LSTM's hidden state, attempts to predict the `next' token. The output predicted token at each step is compared with the `next' ground-truth token, under a cross-entropy loss using the paradigm of Teacher-Forcing ~\cite{teacher_forcing}. To find a good set of model hyper-parameters (e.g.~$L_2$-weights, dropout-rate and \# of LSTM neurons) and the optimal (early) stopping epoch, we use a held-out validation set from Affection and select the model whose generations minimize the
negative-log-likelihood against the ground-truth.

\paragraph{Emotion grounded speaker.}
\label{subsection:emotion_grounded_speaker}

Additionally, following the strategy of affective neural speakers proposed in ArtEmis~\cite{achlioptas2021artemis},
we tested speaking variants trained with Affection that ground their generation with an additional argument aside from the underlying image. Namely, for these \textit{emotion-grounded} variants, during training, we also input to the speaker, at each time-step, an MLP-encoded vector representing the \textit{emotion} the ground-truth explanation justifies. During inference, we
replace the ground-truth emotion with the most likely predicted emotion of the $C_{emotion|image}$ network described above. Interestingly, we observe that this variant also gives, to a first approximation, control over the subjectivity associated with the \textit{individual} reacting to the image, e.g., do they like ‘risky’ activities such as riding a Ferris wheel, or are they afraid of high altitudes? See Figure~\ref{fig:neural_productions_with_emo_grounding} for a demonstration of this induced control resulting from modulating the grounding emotion of this speaking variant.

\paragraph{Pragmatic variants.}
As we show with the experiments in Section~\ref{sec:experimental_results}, the above speakers can already generate plausible explanations supporting a variety of possible emotions for the underlying visual stimulus. To further test the degree to which we can control their output generations (e.g., by including in their output discriminative details of the underlying image), we experiment with their \textit{pragmatic} versions. Specifically, and inspired by the recursive social reasoning characteristic of human pragmatic language usage~\cite{GOODMAN2016818}, we augment the previously described variants with the capacity to prioritize sampled explanations that are deemed to be discriminative, as judged by a separately trained `internal' listener (we use a pretrained CLIP in our experiments). In this case, we sample explanations from our speakers but score (i.e.,~re-rank) them according to:
\begin{equation}
\beta \log(P_L(i,u)) + (1-\beta)\log(P_S(u | i)),
\label{eq:speaker_scoring}
\end{equation}
where $P_L$ is the listener's probability to associate the image ($i$) with a given output utterance ($u$), and $P_S$ is the likelihood of the non-pragmatic speaker version to generate $u$. The parameter $\beta$ controls the relative importance of two terms. We note that because $P_S$ is a probability estimate for a long sequence (generated tokens) evaluated by the product rule, typically, it is orders of magnitude smaller than $P_L$. To address this issue and have human-friendly sets of $\beta$ values in
our ablations, before we apply the above equation, we first re-scale the two probabilities so that, on average, across a test
image set, the two terms to be equal.


\begin{figure*}[htbp]
\centering
\includegraphics[width=\textwidth]{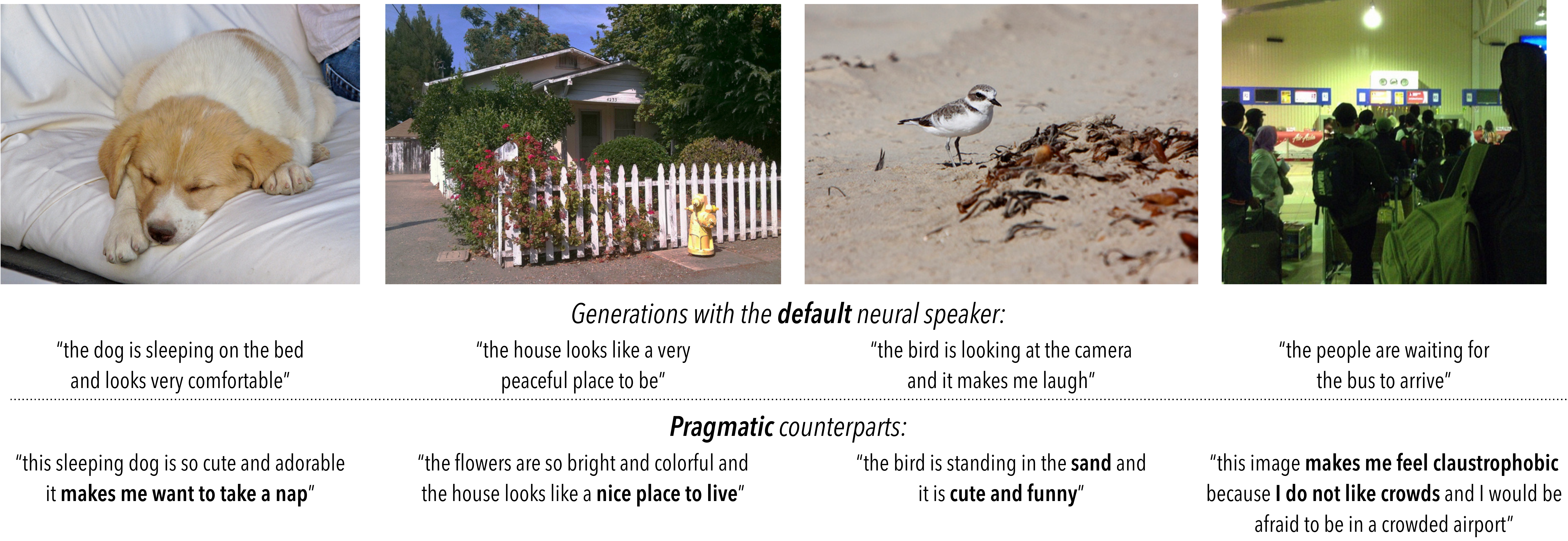}
\caption{\textbf{Effect of boosting the pragmatic content of neural speaker generations via CLIP.} Aside from often correcting the identity of shown objects/actions (right-most image is indeed taken inside an airport), the pragmatic variant tends to use more visual details in its explanations (e.g., \textit{`standing in the sand'}), and perhaps more importantly to expand the explanation to include non-visual but valid associations (e.g., \textit{`take a nap'}, or \textit{`do not like crowds'}). The shown images are included in the datasets covered by Affection, no ownership or copyright claim is made by the authors of this article.}
\label{fig:effect_of_pragmatics}
\end{figure*}

\section{Evaluation}
\label{section:evaluation}

Automatic evaluation of captioning systems remains a challenging problem~\cite{yi-etal-2020-improving, diverse_captions,hessel2021clipscore}. In essence, a captioning system tries to solve a `1-to-many' optimization problem with a handful of soft constraints; i.e., a training image typically admits many \textit{valid} captions even if it is annotated only a few times. Furthermore, captioning systems can (and often do) produce repetitive and overly simple captions across different images, ignoring essential visual cues effectively exhibiting mode collapse~\cite{Shetty2017SpeakingTS,diverse_captions,Seo2020ReinforcingAI}. Both of these problems are exacerbated in our task of Affective Explanation Captioning (AEC) due to its more open-ended and subjective nature, compared to descriptive caption generation. In this work, we do not fully tackle these problems for AEC, as we believe that new and specialized evaluation metrics deserve a separate dedicated treatment. Nevertheless we highlight these challenges in the context of AEC and explore the efficacy and limitations of a large variety of established metrics in relation to different neural speaking variants, providing several key insights.

\paragraph{Comparing with ground-truth.} To evaluate the quality of the output of our neural speakers with respect to the hidden ground-truth annotations of the test images, we first use some of the most established automatic metrics for this purpose: BLEU 1-4~\cite{BLEU}, ROUGE-L~\cite{lin2004rouge}, METEOR~\cite{denkowski:lavie:meteor-wmt:2014} and SPICE~\cite{spice}. These n-gram similarity based metrics, (or semantic-scene-graph-based for SPICE); expect at least \textit{one} of the ground-truth captions to be similar to the corresponding generation. It is worth noting that we do not use CIDEr~\cite{cider}, because unlike the above metrics, CIDEr requires the output generation to be similar to \textit{all} held-out utterances of an image, which by the nature of Affection is not a well-posed desideratum (see Achlioptas \textit{et al.}~\cite{achlioptas2021artemis} for a relevant discussion on this point). 

Aside from the above metrics, which enable comparison of our work with the majority of existing captioning approaches and datasets~\cite{Stefanini2022FromST}, we also explore the utility of the recently proposed metrics CLIPScore and RefClipScore~\cite{hessel2021clipscore}. These metrics leverage the capacity of a pretrained CLIP model to measure the association of an arbitrary caption with a given image and have shown improved correlation with human judgement~\cite{hessel2021clipscore}. Specifically, RefClipScore assumes access to 
ground-truth human annotations which it then compares with the generated caption based on CLIP's association scores. CLIPScore, on the other hand, lifts this latter assumption by directly and solely relying on CLIP's caption-image predicted compatibility, making it a \textit{reference-free} metric. We believe this type of independence from ground-truth annotations that CLIPScore advocates for, is a very promising general direction for evaluating open-ended captions and datasets, like ours.

\paragraph{Assessing diversity of productions.}
To evaluate the susceptibility of different neural speakers to suffer from mode collapse, we consider three metrics. First, we report the average of the \textit{maximum length of the Longest Common Subsequence} (LCS) of the generated captions and (a subsampled version) of all \textit{training} explanations. The smaller the LCS is, the less similar the evaluated captions are from the training data (i.e., less over-fitting occurs). By reporting the maximum found LCS we also dismiss shorter common subsequences, which are inevitable, but which do not contribute to higher scores~\cite{fan2019strategies}. Secondly, we also report for all our neural variants, the percent of \textit{unique} captions they generated across the same set of test images. Last, inspired by the diversity-metric-oriented work of Qingzhong and Antoni~\cite{diverse_captions} and 
the above-described CLIP's application on caption evaluation~\cite{hessel2021clipscore}, we also introduce a new metric, which uses CLIP to detect the lack of caption diversity, and which we dub CLIP-Diversity-Cosine (ClipDivCos)$ \in [-1, 1]$. Specifically for this metric, we use a pretrained CLIP model to encode \textit{all} generated captions across a set of test images, and report the average pairwise \textit{cosine} of the angles of the corresponding embedded vectors. Note that CLIP's textual (and visual) embeddings are optimized to be semantically similar when their angles' cosine is large (+1). Thus, \textit{the smaller} the CLIPDivCos of a collection of vectors is, \textit{the more semantically heterogeneous and diverse} this collection is expected to be.

\paragraph{Specializing to affective explanations.}
The last axis of evaluation that we explore relates to the idiosyncratic properties of \textit{affective explanations}, and for which relevant metrics where introduced in ArtEmis \cite{achlioptas2021artemis}. Concretely, first, we estimate the fraction of a speaker's productions that contain \textit{metaphors and similes}. We do this by tagging generations that include a small set of manually curated phrases like: `thinking of', `looks like' etc. We note that the estimated fraction of \textit{ground-truth} Affection explanations that appear to have such metaphorical-like content is \datasetPrcOfSimiles{} -- setting the default `ideal' and expected percentage for our neural based variants. The second metric introduced in ArtEmis \cite{achlioptas2021artemis}, which we apply to our generations is that of \textit{emotional-alignment}. Specifically, we use the trained $C_{emotion|text}$ classifier to predict the most likely emotion for each generated utterance. This allows us to measure how well this predicted emotion is `aligned' with some pre-defined ground-truth. Namely, for test images where the human-indicated emotions formed a strong majority among their annotators, we report the percent of their corresponding neural based captions where the $\argmax(C_{emotion|caption})$ is equal to the emotion indicated by the majority.

\paragraph{Human-based evaluation.} All previously defined metrics can be automatically computed at scale, and aim to be a proxy for the more expensive and precise human-based quality-assessment of captioning systems~\cite{kilickaya2016re,cui2018learning,hessel2021clipscore}. As a final quality check for our neural based productions, we deploy user-based studies that emulate \textit{emotional Turing tests}~\cite{achlioptas2021artemis}; i.e., they try to assess how likely it is for third-party observing humans to confuse the synthetic captions as if they were made by other humans, instead of being produced by neural speakers.


\begin{figure*}[htbp]
\centering
\includegraphics[width=\textwidth]{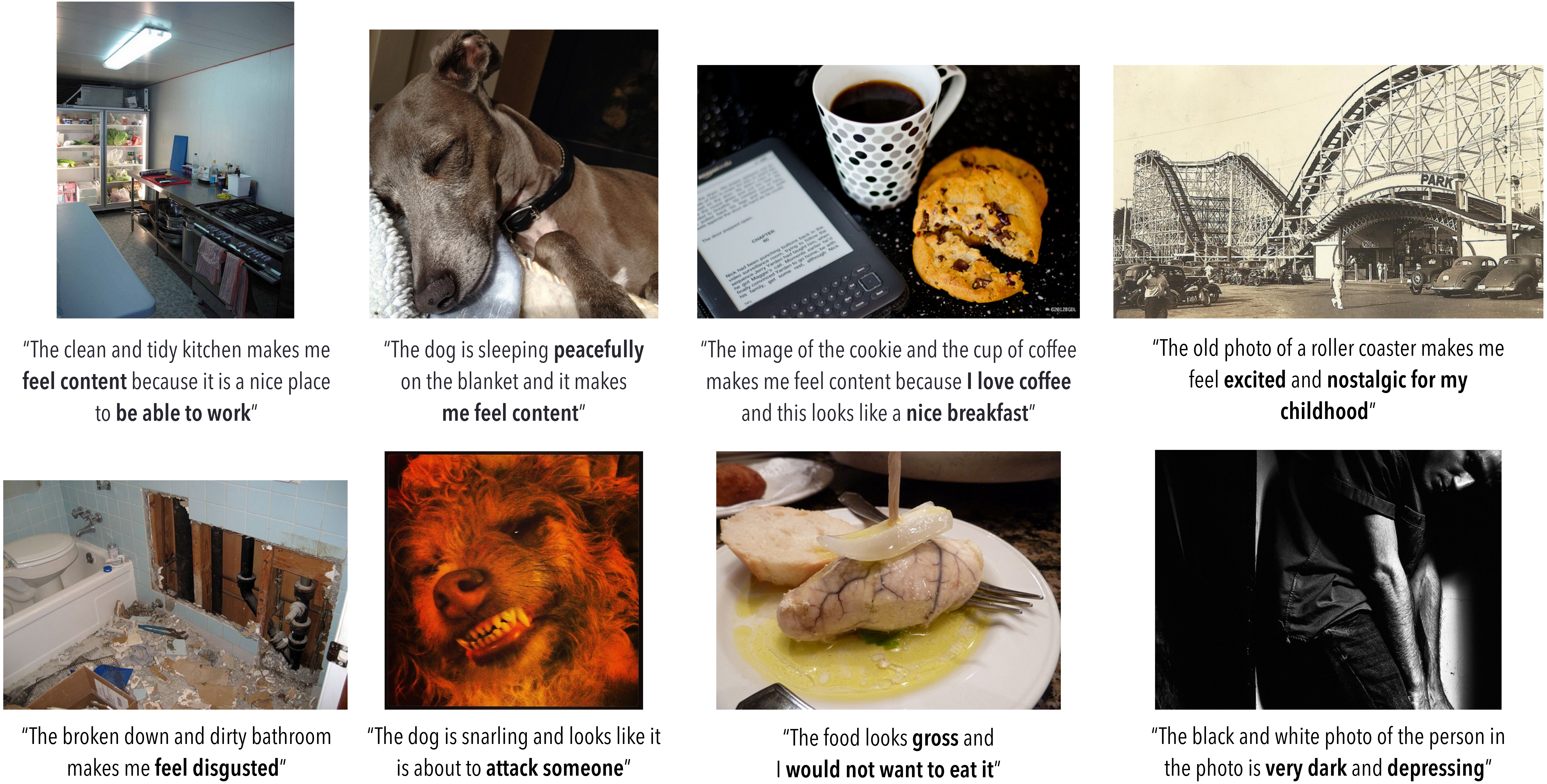}
\caption{\textbf{Curated examples of neural speaker generations on unseen images from the emotion-grounded, pragmatic speaker variant.} The top row includes generations that reflect a positive sentiment, while the bottom row showcases generations grounded on similar visual subjects (object classes) e.g., another dog, food item, etc., that give rise to negative emotions. Remarkably, this neural speaker appears to take into account the underlying fine-grained visual differences to properly modulate its output, providing strong \textbf{explanatory power} behind the emotional reactions. Note, also, how the explanations can include purely human-centric semantics (\textit{`nostalgic of my childhood'}, \textit{`love coffee'}), and use explicit \textbf{psychological} assessments (\textit{`feel content/excited/disgusted'}, \textit{`is depressing'}). The shown images are included in the datasets covered by Affection, no ownership or copyright claim is made by the authors of this article.} 
\label{fig:neural_productions_emo_grounded_pragmatic}
\vspace{-5pt}
\end{figure*}
\section{Experimental Results}
\label{sec:experimental_results}
For the experiments described in this section we train neural networks by using an 85\%-5\%-10\% train/val/test split of Affection, making sure that the splits have no overlap in terms of their underlying images. Moreover, we ignore explanations that contain more than 51 tokens (99-th percentile of token-length across Affection), or those containing fewer than 5 tokens (in total these two constraints remove $\sim 1\%$ of all utterances). Tokens, appearing less than twice in the training set were replaced with a special $\text{<unk>}$ token denoting an out-of-vocabulary word. For more details on our data preprocessing and neural network deployment please refer to the \NameOfAppendix{}~\cite{affection_supp} and the released code. 

\paragraph{Preliminary experiments: emotion classification from text or images.}\label{paragraph:preliminary-experiments}
As shown by previous studies~\cite{achlioptas2021artemis} predicting the \textit{fine-grained} emotion supported by an affective explanation is a much harder task than binary or ternary text-based sentiment classification~\cite{sentiment_analysis_survey,BIRJALI2021107134}. By using the neural-based text predictors described in Section~\ref{subsection:aux_classifiers} we found that an LSTM-based classifier attains $69.8\%$ average accuracy
on the same test split used for our neural-speakers (\numTestUtterances{} explanations). Moreover, a transformer-based (BERT) classifier achieved an improved accuracy of 72.5\% when fine-tuned on this task.  Interestingly, these two models, when trained with ArtEmis \cite{achlioptas2021artemis} generalized more poorly ($63.3\%$ and $64.8\%$, respectively) -- suggesting that the explanations of Affection are more indicative of the the emotion they support, compared to those of ArtEmis. Importantly, these classifiers failed \textit{gracefully} and were mostly confused among subclasses of the same, positive or negative emotion (for confusion matrices please see in the \NameOfAppendix{}). Specifically, if we binarize their output predictions for the above 9-way posed problem along with the ground-truth labels into positive vs.~negative sentiments (ignoring the something-else category); the LSTM-based, and the transformer-BERT-based models, achieve $94.0\%$, $95.5\%$ accuracy, respectively.

For the other relevant problem described in Section~\ref{subsection:aux_classifiers}, that of \textit{image}-to-emotion classification; for the subset of 5,672 test images for which our annotators indicated an emotion that attained a \textit{unique strong majority}, a fine-tuned ResNet-101 encoder predicts the corresponding fine-grained emotion correctly 59.1\% of the time. Notice that, crucially, the emotion label distribution of Affection is highly imbalanced, i.e., across the entire Affection
the smallest emotion group that attains such unique annotation majorities, across all images, is \textit{anger} (.35\%) vs. the largest one is \textit{contentment} (34.9\%). This fact makes our particular prediction problem harder than usual but also open to possibly more specialized solutions such as using a focal loss~\cite{focal_loss,LTVRR}, or curating and using explicitly more balanced emotion labels~\cite{artemis_V2}. Last, and similar to what we observed with the text-to-emotion predictions (above paragraph), if we convert the ResNet predictions to binary-based sentiment ones, the ResNet predicts the ground-truth sentiment 88.5\% of the time correctly, also indicating a graceful failure mode.

\begin{figure}[ht!]
\centering
  \includegraphics[scale=0.55]{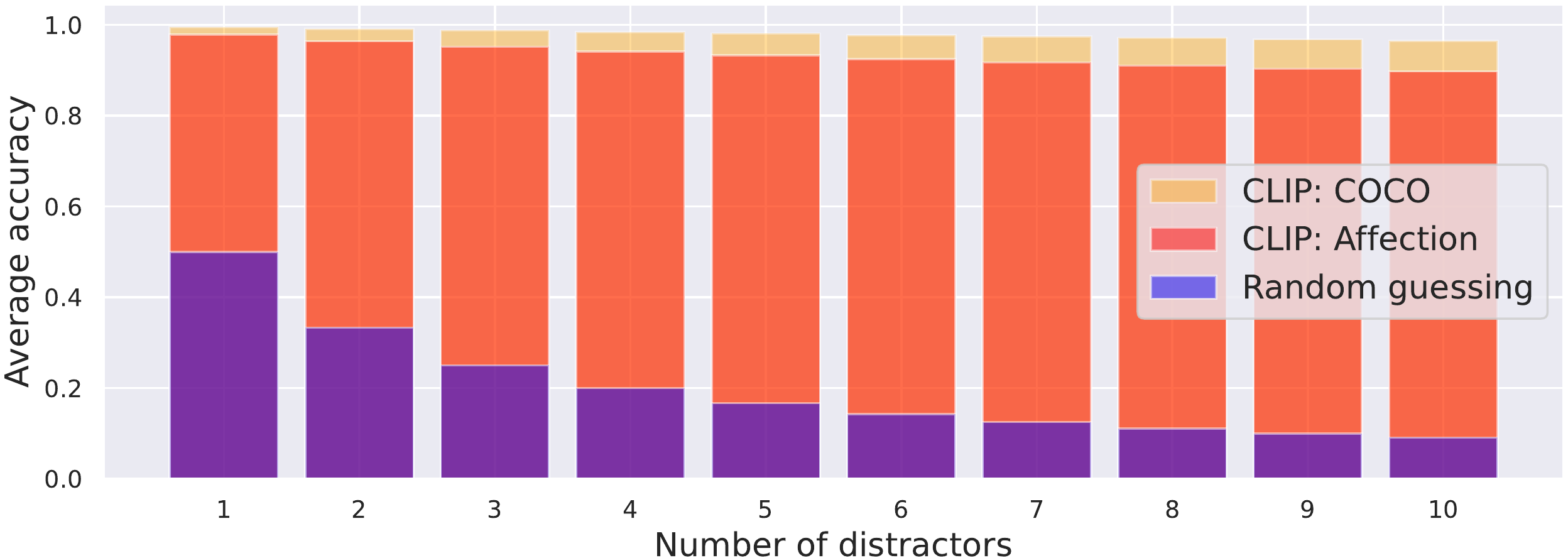}
  \caption{\textbf{Listening accuracy of a pretrained CLIP on the entire collection of Affection and COCO as a function of the number of distracting images used at inference time}. The x-axis displays the number of distracting images, and the y-axis the average accuracy of identifying the corresponding image given a ground-truth caption from either datasets. Random guessing reflects performance when selecting the target uniformly at random. Surprisingly, it appears that Affection contains explanations that describe salient visual elements regarding the image content, to enable \textit{excellent} identification of them, i.e., comparably to the performance of using a purely objective dataset such as COCO.}
  \label{fig:clip_on_affection_and_coco}
\end{figure}

\paragraph{Neural Comprehension of Affective Explanations.}
Next, we explore the extent to which the textual explanations in Affection refer to discriminative visual elements of their underlying images, to enable their \textit{identification} among arbitrary images. Specifically, to answer this question we use the two `neural listeners' described in paragraph~\ref{subsection:paragraph:neural-listeners}. For the CLIP-based experiments we use a pretrained CLIP model with 400M parameters (version \texttt{ViT-B/32}). During inference, we couple all ground-truth image-caption/explanation pairs of a dataset with a varying number of uniformly randomly chosen images from the same dataset -- and upon embedding them in CLIP's joint visio-linguistic space we retrieve for each given caption the image with the largest (cosine-based) similarity.  As can be seen in the results of the average retrieval accuracy displayed in Fig.~\ref{fig:clip_on_affection_and_coco}, Affection's explanations contain significant amounts of `objective' and discriminative grounding details to enable \textit{excellent} identification of an image from its underlying explanation. Specifically, the average accuracy when contrasting the ground-truth pair with a single distracting image is in the very high nineties for both datasets (COCO: 99.5\% vs. Affection: 97.9\%). Moreover, even with as many as ten distracting images the retrieval accuracy remains strong (COCO: 96.5\% vs. Affection: 89.7\%). Interestingly, for either dataset, the drop in performance with the addition of more distracting images is robust (less steep drop than guessing uniformly at random). Finally, we note that the training set of CLIP includes web-scale internet-crawled data,  which are expected to be closer to COCO's nature than to Affection's i.e., affective explanations are possibly not as common online as descriptive image captions -- potentially explaining some of the performance gap observed among the two datasets. 

Regarding the corresponding experiments made with our custom-made, trained with Affection neural listener, we point the interested reader to the \NameOfAppendix{} and note that that variant also showed promising performance in the image retrieval task (e.g., 91.1\% with 1 distractor, and 63.4\% with 10 distractor images), but which was inferior to that attained and shown above with CLIP. More fruitful ways of training affective neural listeners in combination with pre-trained large models (like CLIP), or by jointly training with descriptive captions, is an interesting direction for the future.

\paragraph{Neural-based \underline{A}ffective \underline{E}xplanation \underline{C}aptioning (AEC).}
In our final set of experiments, we deploy the neural speakers described in Section~\ref{subsection:neural-listeners-and-speakers} to give the first neural-based solution to our core problem of AEC. Specifically, we train a basic SAT neural speaker (also denoted as `default'). We also train an emotionally-grounded variant of this model utilizing during inference the fine-tuned ResNet-101 (see the third paragraph of this section); along with emotionally-grounded \textit{and} pragmatic variants that use a pretrained CLIP model (version \texttt{ViT-B/32}) to rank and output the most fitting generation per Eq.~\ref{eq:speaker_scoring}. We evaluate the test generations of these models by applying all metrics described in Section~\ref{section:evaluation}.

Table~\ref{table:speaker_metrics} reports machine-based evaluation metrics and provides several important insights for each variant.  Namely, in this
table, we observe that the old but still widely used primarily n-gram-based metrics of the first group: BLEU-1-4,..., and SPICE, are \textit{slightly} improved if we use the default or default (pragmatic) models, which do not explicitly use emotion for grounding. Given that the held-out explanations typically justify a large variety of emotions for each image, biasing the generation with a single specific (guessed) emotion, as the emotion-grounded variants do, might be too restrictive. On the other hand, for the subset of test images with a ground-truth strong-emotional majority, where we evaluate the emotional-alignment score; we see a noticeable improvement when using the emotion-grounded variants. Interestingly, these variants also fare better regarding the number of similes they produce by better approaching Affection’s ground-truth average of  \datasetPrcOfSimiles.

Regarding the newer metrics of RefClipScore and ClipScore, which have shown a better correlation with human judgment~\cite{hessel2021clipscore}, the pragmatic variants \textit{significantly} outperform their non-pragmatic counterparts. Given that we use CLIP to re-rank and select their generations, this result might be somewhat expected. However, as we will show next (emotional Turing test), these variants also fare better in our competitive human-based evaluation. Also, equally important, for \textit{all} diversity-oriented metrics (third group of metrics in Table~\ref{table:speaker_metrics}), the pragmatic variants fare best. 
For a qualitative demonstration of pragmatic inference's effect see Figure~\ref{fig:effect_of_pragmatics}. Also, for curated generations from the (emotion-grounded) pragmatic variant, please see Figure~\ref{fig:neural_productions_emo_grounded_pragmatic}.


\begin{table}[t]
    \centering
    \resizebox{\linewidth}{!}{
    \begin{tabular}{c c c c c c}        
        \toprule
        \multirow{2}{*}{\textbf{Metrics}} &
        \multicolumn{4}{c}{\textit{Speaker Variants}}  \\
        & \textbf{Default} 
        & \textbf{Emo-Grounded}
        & \multicolumn{1}{p{4cm}}{\centering{\textbf{Default \\ (Pragmatic)}}} 
        & \multicolumn{1}{p{4cm}}{\centering{\textbf{Emo-Grounded \\ (Pragmatic)}}}
        & \multicolumn{1}{p{4cm}}{\centering{\textit{Best \\ Strategy}}} \\
        \midrule
        BLEU-1 ($\uparrow$)             &            \textbf{\cg{64.4}} &                 63.1 &            \cg{64.3} &                 63.4 & \multirow{7}{*}{
      default architecture
        }\\
        BLEU-2 ($\uparrow$)            &            \textbf{\cg{38.3}} &                 36.9 &            \cg{38.0} &                 37.0 &\\
        BLEU-3 ($\uparrow$)            &            \textbf{\cg{22.2}} &                 20.9 &            \cg{21.8} &                 20.9 &\\
        BLEU-4 ($\uparrow$)            &            \textbf{\cg{13.2}} &                 12.0 &            \cg{12.8} &                 11.9 & \\
        METEOR ($\uparrow$)            &            \cg{14.9} &                 14.4 &            \textbf{\cg{15.1}} &                 14.8 &\\
        ROUGE-L ($\uparrow$)           &            \cg{30.8} &                 30.5 &            \textbf{\cg{31.0}} &                 30.8 &\\
        SPICE ($\uparrow$)             &             \cg{7.4} &                  7.2 &            \textbf{\cg{8.0}}  &                  7.7 &\\
        \midrule
        CLIPScore ($\uparrow$)         &            66.7 &                 66.8 &            \textbf{\cg{69.2}} &                 \textbf{\cg{69.2}} &\multirow{2}{*}{pragmatic}\\
        RefCLIPScore ($\uparrow$)       &            75.0 &                 75.0 &            \textbf{\cg{76.3}} &                 \textbf{\cg{76.3}} &\\
        \midrule      
        Unique-Productions ($\uparrow$) &            78.7 &                 80.7 &            \cg{82.9} &                 \textbf{\cg{83.7}} & \multirow{3}{*}{pragmatic}\\
        Max-LCS ($\downarrow$)    &            70.4 &                 70.4 &            \cg{68.6} &                 \textbf{\cg{68.4}} &\\
        CLIP-Cosine ($\downarrow$) &            73.1 &                 72.8 &            \textbf{\cg{69.8}} &                 \cg{70.2} &\\
        \midrule
        Similes ($\downarrow$)  &            42.8 &                 \cg{36.3} &            40.0 &                  \textbf{\cg{34.5}} & 
        \multirow{2}{*}{
      emo-grounded architecture
        }\\
        Emo-Alignment ($\uparrow$) &            48.1 &                 \cg{55.2} &            48.2 &                 \textbf{\cg{55.9}} &\\
        \bottomrule
    \end{tabular}
   }
    \vspace{5pt}
    \caption{\small\textbf{Neural speaker machine-based evaluations}. The Default models use for grounding only the underlying image, while the Emo-Grounded variants also input an emotion-label. Pragmatic variants use CLIP to calibrate the score of sampled productions before selecting the final proposal.}
    \label{table:speaker_metrics}
    \vspace{-10pt}
\end{table}

    
        


\paragraph{Remark.} For all shown qualitative neural speaking results and the results in Table~\ref{table:speaker_metrics}, we remark that our neural speakers are sensitive to the choices we make during \textit{inference} for i) the speaker's (soft-max) temperature, ii) the beam-size of the beam-search sampling algorithm, and iii) the relative importance we assign between the listening vs. speaking compatibility in the pragmatic variants. However, the trends these hyper-parameters create w.r.t. the machine-based evaluation metrics and specifically regarding the `Best Strategy' (Table~\ref{table:speaker_metrics}) one should follow to maximize each metric, are very stable and predictable~\cite{vedantam2017context,achlioptas2019shapeglot,achlioptas2021artemis}. 

\paragraph{Emotional Turing test.} 
With this test, we evaluate how likely our neural speaking variants' output generations can be perceived as if they were made by humans. Specifically,  we first form a random sample of 500 \textit{test} images and accompany each image with one of their ground-truth, human-made explanations. We then couple each such image/explanation with a generation made by a neural speaker. We do this by considering all our four speaking variants to obtain 2,000 image-caption samples (an image paired with two explanations, a `neural-' and a `human-' based one).  We then proceed by asking AMT annotators who have never seen these sampled images to observe them and, upon reading closely
the coupled explanations, to select one among four options indicating that: (a) \textit{both} explanations seem to have been made by humans justifying their emotional reaction to the shown image; (b) \textit{none} of the explanations are likely to have been made by humans for that purpose, or (c) (and (d)) to select the explanation that seems more likely to have been made by a human. The findings of this emotional Turing test are summarized in Figure~
\ref{fig:emo_turing_results}. As can seen in this figure, for all variants, more than 40\% of the time (41.4\%-46.2\%), both displayed utterances were thought as if humans made them (blue bars). Moreover, and perhaps somewhat surprisingly at first reading, in a significant fraction of the answers, the neural-based generations were deemed more likely/fitting than the human-made ones (green bars of the same figure). These results highlight both the complexity of the AEC problem as well as the promising overall quality of our neural speaker solutions, enabled by the Affection dataset. 

\begin{figure*}[ht]
    \centering
  \includegraphics[scale=0.55]{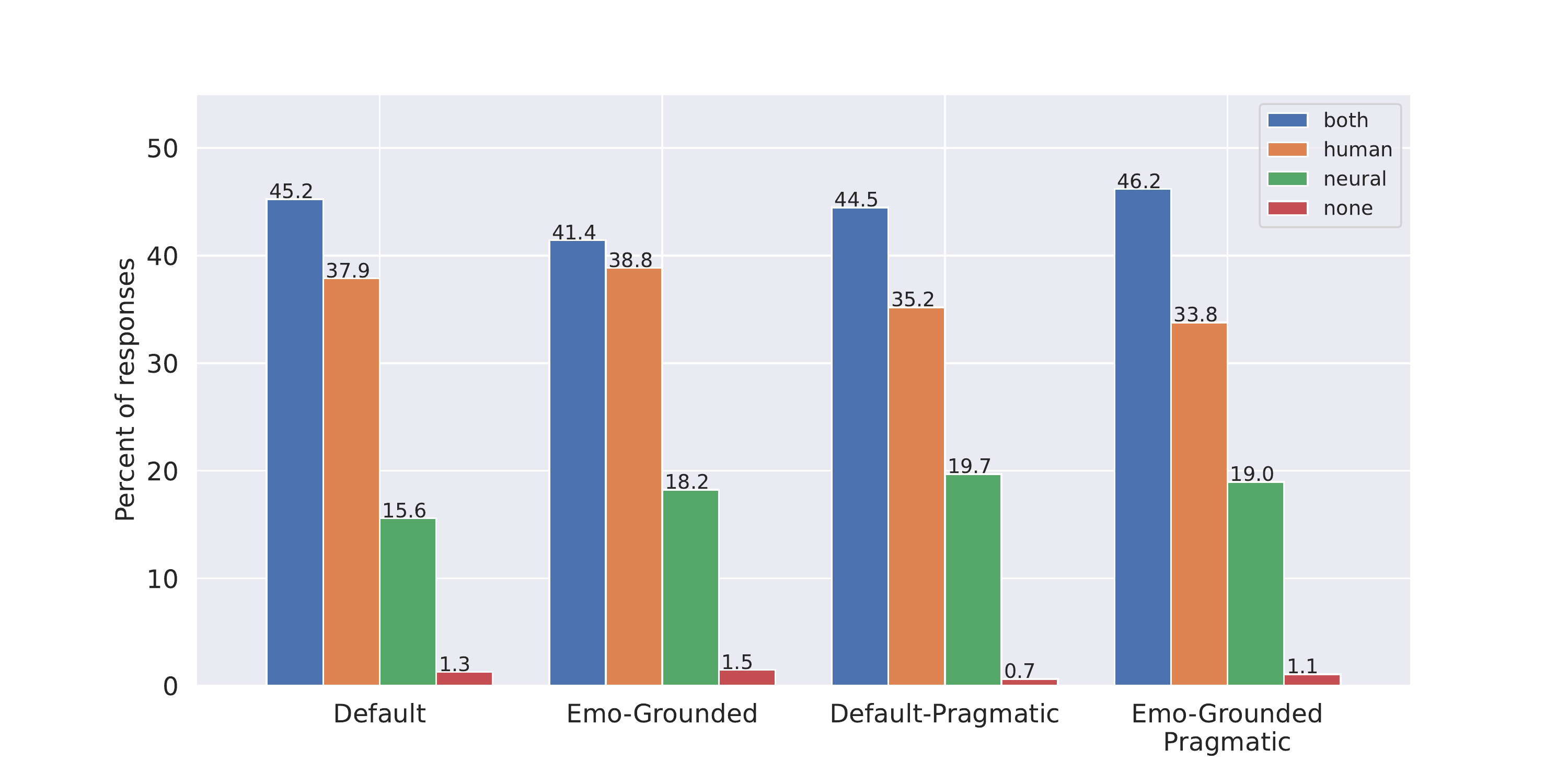}
  \caption{{\bf Turing test results for our neural speaking variants.} For each variant we show the percent of its evaluated explanations that fall in one of the four categories ("both", "human", "neural" and "none") described in Section~\ref{sec:experimental_results}. All variants show strong baseline performance, with a minimum \textit{aggregate} success rate ("both" and "neural" bars) of 59.6\% attained by the emo-grounded variant, and a maximum rate of 65.2\% attained by its (emo-grounded) pragmatic version. Note, that both pragmatic variants (two right-most histograms) outperformed their non-pragmatic versions.}
  \label{fig:emo_turing_results}
\end{figure*}

\section{Conclusion and Future Vision}

Humans react to stimuli beyond the literal content of an image as they respond to the story behind it. More broadly the image evokes emotions relating to human experience in general and the viewer’s experience in particular. In this work we have shown how linguistic explanations of affective responses can express and illuminate this larger context, leading to a more comprehensive understanding of how image content and its elements affect human emotion. Using our new Affection dataset, we have demonstrated that AI-powered neural speakers can produce generations that mimic well these human responses. In addition to highlighting the importance and utility of exploring the affective dimension of image understanding, we believe that this study can stimulate work in many interesting novel directions. For example, exploring this richer scope of responses to visual stimuli will be essential in creating AI-powered assistants, whether virtual or embodied that can interact with humans \textit{in seamless ways}, as they can better \textit{bond} with users by responding to more nuanced subjective dimensions of the visual experience.

\section*{Acknowledgments}
P.A. wants to thank Professors James Gross, Noah Goodman and Dan Jurafsky for their initial discussions and the ample motivation they provided for exploring this research direction. Also, wants to thank Ashish Mehta for fruitful discussions on alexithymia~\cite{alexithymia}. Last but not least, the authors want to emphasize their gratitude to all the hard-working Amazon Mechanical Turkers without whom this work would be impossible. 

Parts of this work were supported by the ERC Starting Grant No. 758800 (EXPROTEA), the ANR AI Chair AIGRETTE, and a Vannevar Bush Faculty Fellowship.

}

\ifthenelse{\equal{\paper}{supp}}{
\appendix
\section{Details on Building \datasetName{}}
The \datasetName{} is utilizing images taken from the following {\bf five} publicly available datasets:
    \begin{enumerate}
    \item \nospacepara{`Yang-Affective'} All 21,829 images used in the study of~\cite{emotion_clf} and provided at \href{https://onedrive.live.com/redir?resid=AB6522E29F6ED9A0!101730&authkey=!AH57YMUbsP-qNls&e=u2Tv7I}{link} found in the the webpage of \href{https://qzyou.github.io/}{Quanzeng You}. These images are coupled with DESC-based labels classifying each image in one of the eight emotional states used in \datasetName{}.
    
    \item  \nospacepara{`Emotional-Machines-Affective'} All \textbf{10,767} images used in the study of~\cite{emotion_clf_valence_arousal} downloaded from \href{https://figshare.com/articles/dataset/CGnA10766_Dataset/5383105?file=9263326}{link-to-download}. These images are coupled with Valence-Arousal affective labels curated in~\cite{emotion_clf_valence_arousal}.
    
    \item \nospacepara{`COCO-Affective'} 22,770 images from the caption-oriented data set of COCO~\cite{coco_chen2015} .
    
    \item \nospacepara{`FlickR30K-Affective'} 13,202 images from the caption-oriented data set of \flickr~\cite{plummer2015flickr30k}.
    
    \item \nospacepara{`Visual-Genome-Affective'} 16,437 images from the data set of Visual Genome~\cite{krishna2017visual}.
\end{enumerate}

For these last three (descriptive) data sets we annotated with affective explanations a subset of their images that were semantically similar to the 21,829 images of~\cite{emotion_clf}. Specifically, we included an image of such a data set in \datasetName{} if it was among the three L2-based nearest-neighbors of any image of~\cite{emotion_clf}. To capture semantic proximity and collect these neighbors we used a 512D embedding vector space formed by the output weights of the final convolutional layer of a ResNet-32~\cite{resnet} pre-trained on ImageNet~\cite{deng2009imagenet}. The $7 \times 7$ spatial dimensions of the output of this layer were pooled together by applying average pooling across them (forming a $1 \times 1 \times  512$ embedding vector per image). For the Visual Genome, it is worth noting that we we downloaded all of (108,077) images from the \href{https://visualgenome.org/api/v0/api_home.html}{its official web page} and restricted the nearest-neighbor search on the 56,506 images that were \textit{not} included in COCO~\cite{coco_chen2015}, or \flickr~\cite{plummer2015flickr30k}.

Upon selecting the aforementioned images from all corresponding five datasets, we used ``fdups''\cite{fdupes}  to discover possible duplicates among them. We found 198 duplicates which were removed from our final version of Affection.

\section{Analyzing more properties of \datasetName{}}

\begin{figure*}[ht]
  \includegraphics[width=\textwidth]{figures/supp_mat/Affection_analysis_per_img_dataset.pdf}
  \caption{{\bf Measuring key properties of {\datasetName} across its underlying image datasets}. Histograms comparing {\datasetName} in each of its underlying image datasets along the axes of (a) \textit{Concreteness}, (b) \textit{Subjectivity}, and (c) \textit{Sentiment}. The explicitly emotion oriented datasets of Emotional-Machines~\cite{emotion_clf_valence_arousal} and of You \textit{et al.}~\cite{emotion_clf} result in only slightly more abstract, subjective and sentimental language, compared to the subset of images used from the remaining, more in-the-wild, datasets.}
  \label{fig:analysis_dataset_per_image_subcollection}
\end{figure*}

\begin{figure*}[ht]
  \includegraphics[width=\textwidth]{figures/supp_mat/Affection_vs_ArtEmis.pdf}
  \caption{{\bf Comparing {\datasetName} to ArtEmis}.}
  \label{fig:comparingAffectionToArtEmis}
\end{figure*}

\begin{figure}[ht]
  \includegraphics[width=\linewidth]{figures/analysis_of_dataset/Affection_wordcloud.png}
  \vspace{-3mm}
  \caption{{\bf Wordcloud with common words (tokens) present in \datasetName}.\vspace{-2mm}}
  \label{fig:dataset_wordcloud}
\end{figure}

\section{Direct Emotion Prediction from a Single Modality}

\subsection{Text-to-Emotion Classification}
\begin{figure}[ht]
  \includegraphics[width=\linewidth]{figures/supp_mat/lstm_text2emo_confusion_matrix.png}
  \vspace{-3mm}
  \caption{{\bf Confusion matrix for an LSTM-based text2emotion 9-way classifier}.}
  \label{fig:lstm_text2emo_confusion_matrix}
\end{figure}

\subsection{Image-to-Emotion Classification}

\subsection{Image-to-Emotion Classification}
In this line of experiments we try the generalization of a pre-trained ResNet-101, finetuned on Affection labels, \textbf{only} on images with strong-emotional majority per their ground-truth annotations. This distribution is highly imbalances as can be seen in Figure~\ref{fig:fraction_of_images_with_strong_emo_majority_hist}.

\begin{figure}[ht]
  \includegraphics[scale=0.44]{figures/supp_mat/fraction_of_images_with_strong_emo_majority_hist.pdf}
  \vspace{-3mm}
  \caption{{\bf Fraction of Affection images that have strong emotional majority in each given emotion class}.}
  \label{fig:fraction_of_images_with_strong_emo_majority_hist}
\end{figure}

\begin{figure}[ht]
  \includegraphics[width=\linewidth]{figures/supp_mat/im2emotion_resnet101_confusion.png}
  \vspace{-3mm}
  \caption{{\bf Confusion matrix for a ResNet-101 pretrained image2emotion 9-way classifier.}.}
  \label{fig:resnet101_img2emo_confusion_matrix}
\end{figure}

\section{Neural Listeners and Speakers based on Affective Captions}

\subsection{Neural Listeners}
\begin{figure}[ht!]
  \includegraphics[scale=0.6]{figures/supp_mat/test_performance_5_seeds_n_distractors.pdf}
  \caption{\textbf{Listening accuracy of an \textit{in-house} trained, with Affection captions, contrastive architecture similar to CLIP}. The performance displayed is a function of the number of distractor images used at inference time and is the average of using five random seeds when pairing the target with randomly selected distractor images. Random guessing reflects performance when selecting the target uniformly at random. As expected, our neural listener fairs significantly better, but also decreases its performance less more smoothly when more distractors are considered.}
\end{figure}

\subsection{Neural Speakers}
\paragraph{Common Failure Modes}
The first problem that is faced by \textit{all} of our neural speakers variants is oftentimes their inability to recognize the underlying object classes of the image depicted objets. In their generations thus they might ground their explanations on objects not actually shown at the input image, e.g., describe properties of a male human when only females are shown. This generic error appears in many captioning systems and is not specific to affective speakers. However, on affective imagery the problem can be more severe, since such data typically have more subtle and abstract semantics (see Fig.\ref{fig:neural_speaker_failure_examples}-(A)).

\begin{figure}[ht]
  \includegraphics[width=\linewidth]{figures/supp_mat/neural_speaker_failure_examples.pdf}
  \caption{\textbf{Common Failing Modes of Affective Neural Speakers.} (A) generation by default speaker architecture, \textit{all} variants demonstrated the same problem of confusing the clock for a pizza. (B)... }
  \label{fig:neural_speaker_failure_examples}
\end{figure}

}

\bibliographystyle{alpha}
\bibliography{references}


\end{document}